\RequirePackage{fix-cm}
\PassOptionsToPackage{table,dvipsnames}{xcolor}
\documentclass{style}
\DeclareMathSizes{13}{12}{9}{7}

\usepackage{wrapfig}
\usepackage{booktabs}
\usepackage{amssymb}
\usepackage{multirow}
\usepackage{subcaption}
\usepackage{tabularx}
\usepackage{array}
\usepackage{makecell}
\usepackage{tikz}
\usepackage[normalem]{ulem}
\usepackage{placeins}
\usepackage{float}

\definecolor{rainbowRed}{HTML}{FF5252}
\definecolor{rainbowOrange}{HTML}{FFB74D}
\definecolor{rainbowGreen}{HTML}{25F0B1}
\definecolor{rainbowBlue}{HTML}{33ABF9}
\definecolor{rainbowIndigo}{HTML}{5467CB}

\definecolor{tplSpecial}{HTML}{6A1B9A}
\definecolor{tplSlot}{HTML}{C2185B}
\definecolor{tplBg}{HTML}{F0F0F0}
\definecolor{tplTitleBg}{HTML}{222222}
\definecolor{tplFrame}{HTML}{222222}
\definecolor{tplText}{HTML}{1A1A1A}
\newcounter{tplbox}

\newcommand{\pcell}[1]{\multicolumn{1}{|c|}{#1}}
\newcommand{\resultarrow}{\textcolor{gray!65}{\(\rightarrow\)}}

\fancypagestyle{firstpage}{
  \fancyhf{}
  \fancyhead[L]{\raisebox{-1.5pt}{\includegraphics[height=1.5cm, width=3.5cm, keepaspectratio]{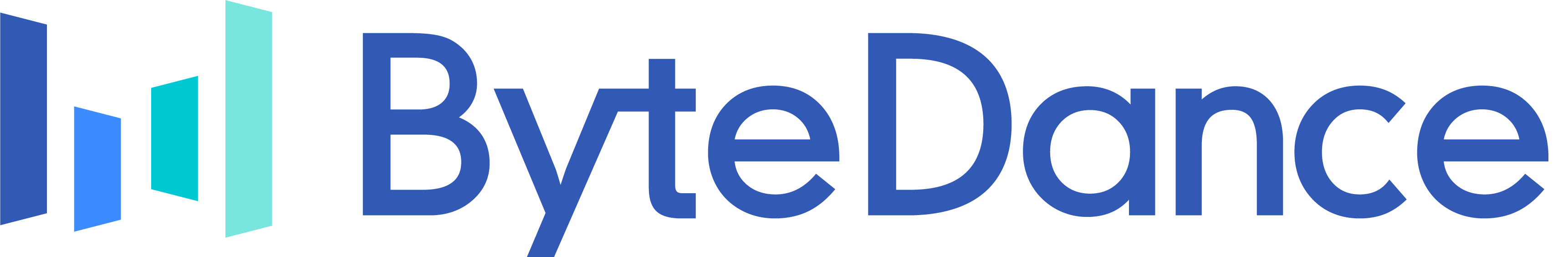}}}
  \fancyfoot{}
  \fancyfoot[C]{\thepage}
  \renewcommand{\headrulewidth}{2.5pt}
}

\title{
    \fontsize{20pt}{24pt}\sffamily\bfseries\color{black}
    \raisebox{-0.22\height}{\includegraphics[height=1.0em, keepaspectratio]{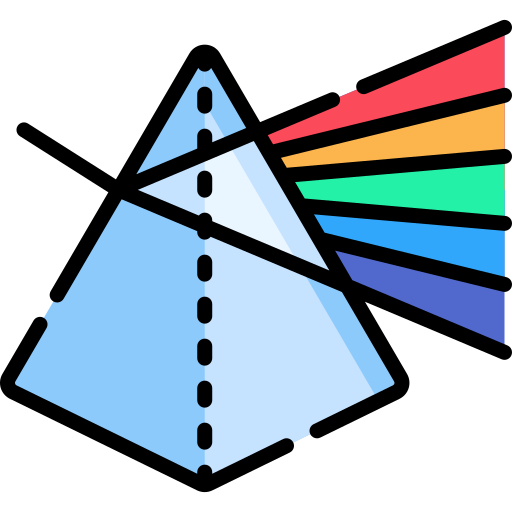}}\hspace{0.1em}%
    PRISM: \textcolor{rainbowRed}{P}riority-aware \textcolor{rainbowOrange}{R}ubric \textcolor{rainbowGreen}{I}nternalization \\ via \textcolor{rainbowBlue}{S}tructured \textcolor{rainbowIndigo}{M}ultimodal Data Synthesis
}
\author{
    \fontsize{13pt}{15.5pt}\selectfont\sffamily\color{black}
    Xiaomin He\textsuperscript{1,2,*,\textdagger} \quad Dongling Xiao\textsuperscript{1,*} \quad Jiahao Xie\textsuperscript{1} \quad Ruiqi Lu\textsuperscript{1} \\[0.3em]
    Qianle Wang\textsuperscript{1} \quad Zhongbin Guo\textsuperscript{1,\textdagger} \quad Wanxuan Sun\textsuperscript{1,\faEnvelope[regular]} \\[0.5em]
    \normalsize\sffamily\color{black} \textsuperscript{1} ByteDance \quad
    \normalsize\sffamily\color{black} \textsuperscript{2} Peking University
}

\email{2401210613@stu.pku.edu.cn}
\hypersetup{
    pdftitle={PRISM: Priority-aware Rubric Internalization via Structured Multimodal Data Synthesis},
    pdfauthor={Xiaomin He, Dongling Xiao, Jiahao Xie, Ruiqi Lu, Qianle Wang, Zhongbin Guo, and Wanxuan Sun}
}

\begin{document}

\maketitle
\thispagestyle{firstpage}
\enlargethispage{1.5\baselineskip}
\makeatletter
\renewcommand{\@fnsymbol}[1]{%
  \ifcase#1\or *\or \textdagger\or \faEnvelope[regular]\else\@ctrerr\fi}
\renewcommand{\@makefntext}[1]{%
  \noindent\makebox[1.5em][c]{\@makefnmark}#1}
\makeatother
\renewcommand{\thefootnote}{\fnsymbol{footnote}}
\footnotetext[1]{Equal contribution.\hspace{1.5em}%
  \makebox[1.5em][c]{\textsuperscript{\faEnvelope[regular]}}Corresponding author.}
\footnotetext[2]{Work done during an internship at ByteDance.}
\renewcommand{\thefootnote}{\arabic{footnote}}
\setcounter{footnote}{0}

\begin{abstract}
Real-world multimodal instructions often bundle multiple requirements with unequal importance, yet most multimodal training data still reduce instruction following to answering one self-contained question. We study this gap through \textbf{rubric comprehension}, which casts the model not as a generator measured against rubrics but as an \textbf{executor} that follows them: given an image and a typed, prioritized rubric, the model must verify each rule before producing an overall judgment. To support this setting, we propose \textbf{PRISM}, a four-stage data synthesis framework that produces persona--task pairs, prefix-guided rule sets, quality-filtered rubrics, and structured verification traces. We further introduce \textbf{PRISM-Eval}, whose Loose and Strict metrics use deterministic matching against fixed labels and therefore require no inference-time judge model. With only 10K synthesized samples, PRISM lifts Qwen3-VL-4B from 9.5\% to 30.1\% Strict accuracy on PRISM-Eval while preserving average performance on general benchmarks, and the gains transfer to four additional open-source MLLMs across dense and MoE architectures, suggesting that structured rubric supervision is a scalable path toward multi-rule, priority-aware multimodal instruction following.
\end{abstract}

\section{Introduction}

Multimodal Large Language Models (MLLMs) have made remarkable progress, approaching or surpassing human-level performance on standard benchmarks of VQA, captioning, and document understanding~\citep{openaiGPT5,baiQwen3VLTechnicalReport2025}. Yet as these models move from controlled benchmarks to real-world deployment, a class of user instructions consistently exposes their limits: instructions whose intent is not a single question, but a bundle of requirements that must be jointly satisfied.

\begin{wrapfigure}[10]{r}{0.5\textwidth}
    \centering
    \vskip -0.6cm
    \includegraphics[width=0.48\textwidth]{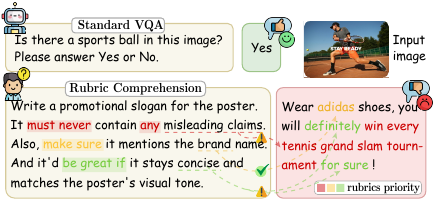}
    \caption{\textbf{From standard VQA to rubric comprehension.} For instance, when asked to write a poster slogan, the user may have multiple requirements.}
    \label{fig:intro}
    \vskip -0.2cm
\end{wrapfigure}
Each such requirement is a rule (Figure~\ref{fig:intro}). Since rules differ in importance, faithfully interpreting the instruction is not a single-question answer: the model must parse every rule and reason about their relative priorities. Such instructions appear in rule-based discriminative tasks (e.g., product quality inspection, advertisement compliance review) and constrained generation. This work focuses on the former, using discriminative verification as a measurable proxy for \textbf{multi-rule, priority-aware comprehension}.

To systematically train and evaluate this capability, we operationalize it as a discriminative proxy task that we call \textbf{rubric comprehension}. Here the user's set of rules is organized into a structured, prioritized collection, a \textbf{rubric}, and the model must produce a judgment for each rule it contains.

The core obstacle is a supervision gap: existing multimodal instruction-tuning datasets~\citep{luo2024mmevol,yao2024mulberry,xu2024llavacot,zhang2025oasis} largely follow the paradigm of ``understand the image and answer a question'' and rarely require interpreting a structured set of prioritized rules. Crucially, scale alone cannot close this gap. Even the strongest closed-source MLLMs remain far from solving rubric comprehension (\S\ref{sec:experiments}), indicating that the bottleneck lies in the \emph{kind} of supervision rather than in model capacity.

Our key insight is a change of perspective. Prior work predominantly places rubrics on the \emph{judgment} side: they serve either as inference-time criteria for benchmark scoring~\citep{zhou2023ifeval,he2025advancedif} or as training-time reward signals~\citep{gunjal2025rar,xu2026rubricarm}. We instead place rubrics on the model-input side and train the model to apply them as an \textbf{executor}: it parses every rule, accounts for priority, and produces both rule-level judgments and an overall verdict. Teaching this executor-side behavior requires supervision that explicitly factorizes an instruction into typed, prioritized rules and demands a verdict on each, precisely what current data lack.

To operationalize this insight, we propose \textbf{PRISM}, a data synthesis framework for teaching MLLMs to \textbf{internalize} multi-rule instructions rather than treat them as one-shot prompts. We first design a structured rubric representation that organizes rules into five types (Perceptual, Reasoning, Content, Format, Linguistic) and three priority levels (Critical, Important, Optional). On top of this, our four-stage pipeline comprises (1) persona--task discovery, (2) prefix-guided rule completion, (3) rule quality assessment, and (4) structured response generation. We pair the framework with \textbf{PRISM-Eval}, a companion evaluation protocol supporting both Loose and Strict rule-based automatic assessment without relying on external judge models.

With only 10K synthesized samples, PRISM substantially outperforms prior data-synthesis baselines on PRISM-Eval and closes much of the gap to the strongest closed-source frontier, while preserving general-benchmark performance. The same gains transfer to four additional MLLMs covering diverse architectures and scales.

Our contributions are summarized as follows:
\begin{itemize}[leftmargin=*,topsep=4pt,itemsep=4pt,parsep=0pt]
    \item We formulate rubric comprehension as an executor-side task and design a fixed-label evaluation protocol (Loose\,/\,Strict accuracy) whose scoring requires no external judge model.
    \item We propose \textbf{PRISM}, a four-stage data synthesis framework that produces typed, priority-aware training signals for rubric comprehension.
    \item PRISM substantially improves in-domain Strict accuracy across five diverse MLLM backbones while preserving general-benchmark performance.
\end{itemize}

\section{Related Work}

\noindent\textbf{Multimodal Data Synthesis.}
Automated production of multimodal instruction data has largely followed two routes. \textbf{Early efforts}~\citep{luo2024mmevol,yao2024mulberry,xu2024llavacot} expand sample complexity and reasoning diversity through evolution, collective Monte-Carlo tree search, or staged ``summary--caption--reasoning--conclusion'' chain-of-thought. These methods all rely on carefully designed prompt templates or handcrafted rules to steer the generation direction. \textbf{OASIS}~\citep{zhang2025oasis}, inspired by Magpie~\citep{xu2024magpie} in the text-only domain, feeds an MLLM only an image and the dialogue-template prefix and lets it freely complete instructions and responses. This removes prompt engineering, but leaves the data distribution entirely determined by the model's own biases. PRISM inherits the simplicity of prefix completion but adds intent-guided prefixes and rule-level quality assessment, steering synthesis toward multi-rule, priority-aware rubric comprehension.

\noindent\textbf{Rubric-Based Methods.}
Rubrics---structured, multi-dimensional evaluation criteria---have been widely adopted in the LLM community, yet almost exclusively on the \textbf{judgment} side. On the evaluation side, IFEval~\citep{zhou2023ifeval}, InFoBench~\citep{qin2024infobench}, and AdvancedIF~\citep{he2025advancedif} decompose complex instructions into programmatically verifiable or fine-grained sub-criteria that serve as scoring yardsticks. On the training side, RIFL~\citep{he2025advancedif}, Rubrics-as-Rewards~\citep{gunjal2025rar}, and Rubric-ARM~\citep{xu2026rubricarm} convert rubrics into multi-dimensional RL rewards, while RuscaRL~\citep{ruscarl2025} further uses rubrics as decaying rollout-time scaffolding. Across these works, rubrics remain on the judgment side; PRISM instead studies the comparatively underexplored setting in which rubrics are supplied as model inputs and jointly supervise rule-level verification and priority-aware aggregation.

\section{Method}

We first formalize the rubric comprehension task and its evaluation protocol (\S\ref{sec:formulation}), then describe our data synthesis pipeline \textbf{PRISM} (\S\ref{sec:synthesis}), and finally instantiate it into the training and evaluation pools together with their composition (\S\ref{sec:dataset}).

\subsection{Problem Formulation}
\label{sec:formulation}

We cast rubric comprehension as a conditional generation task. The input is a pair $(I, \mathcal{R})$, where $I$ is an image and $\mathcal{R} = \{(t_i, p_i, c_i)\}_{i=1}^{N}$ is a rubric of $N$ rules, each rule specified by a type $t_i \in \mathcal{T}$, a priority $p_i \in \{\mathrm{Critical}, \mathrm{Important}, \mathrm{Optional}\}$, and a description $c_i$. The model produces a structured trace $\tau = (g,\, \{(j_i, e_i)\}_{i=1}^{N},\, y)$: visual grounding $g$, per-rule verifications $(j_i, e_i)$ with $j_i \in \{\mathrm{Pass}, \mathrm{Fail}\}$, and aggregated conclusion $y$ that respects the priority hierarchy. PRISM learns $p_\theta(\tau \mid I, \mathcal{R})$ via SFT on synthesized $(I, \mathcal{R}, \tau)$ tuples.

\paragraph{Evaluation Protocol.}
Given a model output $\hat{\tau}$, we measure rubric comprehension at two granularities. Let $y$ and $\hat y$ denote the gold and predicted overall verdicts, and let $j_i$ and $\hat j_i$ denote the corresponding rule-level labels. For a rubric with $N$ rules,
\[
\mathrm{Loose}=\mathbb{1}[\hat y=y],
\qquad
\mathrm{Strict}=\mathbb{1}[\hat y=y]\prod_{i=1}^{N}\mathbb{1}[\hat j_i=j_i].
\]
Strict is our primary metric because it requires the overall verdict and every rule judgment to be correct. ``Judge-free'' refers specifically to scoring: predictions are matched deterministically against fixed labels, without invoking an inference-time LLM judge; it does not assume that synthesized labels are intrinsically error-free.

\subsection{PRISM: Data Synthesis Framework}
\label{sec:synthesis}

\begin{figure*}[t]
  \centering
  \includegraphics[width=1.0\textwidth]{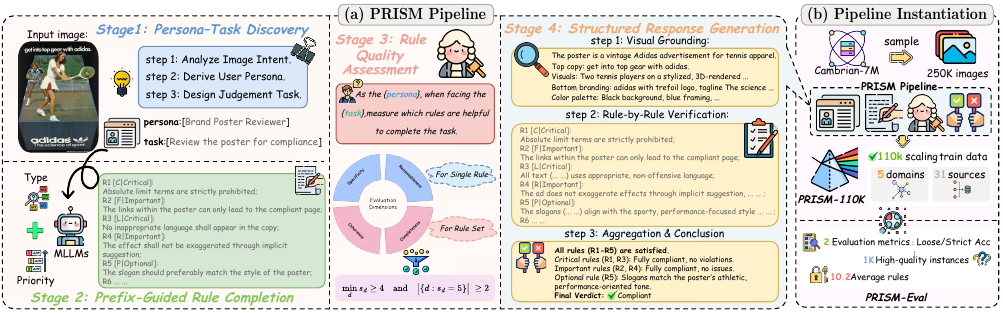}
  \caption{\textbf{Overview of the PRISM data synthesis pipeline.} Part \textbf{(a)} shows the four stages of PRISM: (1)~Persona--Task Discovery, (2)~Prefix-Guided Rule Completion, (3)~Rule Quality Assessment, and (4)~Structured Response Generation, which produces the supervision trace $\tau{=}(g,\{(j_i,e_i)\},y)$. PRISM is employed to generate \textbf{PRISM-110K} for SFT and \textbf{PRISM-Eval} for evaluation, as shown in part \textbf{(b)}.}
  \vspace{-0.4cm}
  \label{fig:pipeline}
\end{figure*}

Given an image $I$ sampled from the Cambrian dataset~\citep{tong2024cambrian}, PRISM synthesizes $(I, \mathcal{R}, \tau)$ tuples through four stages: (1) persona--task discovery, (2) prefix-guided rule completion, (3) rule quality assessment, and (4) structured response generation. We use frozen generators with stage-specific prompts: Seed2.0 Lite~\citep{seed2modelcard} for Stages~1, 3, and 4, and Qwen3-VL-30B~\citep{baiQwen3VLTechnicalReport2025} for Stage~2 rule completion. This separation makes rule generation distinct from downstream filtering and label synthesis. The overall pipeline is shown in Figure~\ref{fig:pipeline}.

\paragraph{Stage 1: Persona--Task Discovery.}
Given image $I$, we prompt Seed2.0 Lite to first extract its \emph{image intent}---the most plausible real-world use case for $I$---and then derive an associated \emph{persona} (a role identity) and \emph{task} (a task description). For instance, a geometry diagram may yield the persona ``middle-school math test-paper proofreader'' and the task ``check whether this geometry diagram meets standard exam requirements''. This anchoring keeps downstream rubrics grounded in plausible deployment scenarios rather than artificial constructs; the full prompt is reproduced in Figure~\ref{fig:persona-prompt}.

\paragraph{Stage 2: Prefix-Guided Rule Completion.}
Conditioned on the persona and task, we construct a structured prefix that encodes our rule-type taxonomy and prompt Qwen3-VL-30B to complete a rubric of multi-dimensional rules. The type space $\mathcal{T}$ used in \S\ref{sec:formulation} is instantiated as five categories:

\begin{itemize}[leftmargin=*,topsep=2pt,itemsep=3pt,parsep=0pt]
    \item \textbf{P (Perceptual):} \textit{object presence, attributes, counts, spatial layout.}
    \item \textbf{R (Reasoning):} \textit{comparison, arithmetic, deduction over visual or symbolic content.}
    \item \textbf{C (Content):} \textit{topical scope and factual coverage of the answer.}
    \item \textbf{F (Format):} \textit{schema, layout, section ordering.}
    \item \textbf{L (Linguistic):} \textit{style, tone, register, terminology.}
\end{itemize}

Each rule additionally carries a priority $p_i$ (Critical / Important / Optional) that reflects its relative importance. Rules are produced sequentially, each emitted with a leading \texttt{[Type|Priority]} tag that drives the generator to span the taxonomy and avoid homogenization. Compared to unconstrained free-form generation, this prefix-guided mechanism markedly improves rule diversity and structural completeness; the exact prefix template is shown in Box~\ref{box:stage2-prefix}.

\refstepcounter{tplbox}\label{box:stage2-prefix}
\begin{tcolorbox}[
    breakable,
    colback=tplBg,
    colframe=tplFrame,
    coltext=tplText,
    coltitle=white,
    colbacktitle=tplTitleBg,
    fonttitle=\bfseries,
    title={Prefix Template},
    boxrule=0.6pt,
    arc=2pt,
    left=6pt, right=6pt, top=4pt, bottom=4pt
]
{\small\ttfamily\sloppy\hyphenpenalty=10000\exhyphenpenalty=10000
\textcolor{blue}{<|im\_start|>}user\\
\textcolor{blue}{<|vision\_start|><|image\_pad|><|vision\_end|>}\\[2pt]
Suppose you are a \textcolor{red}{\{persona\}}, \textcolor{red}{\{task\}}. Please make your judgment based on the following rubrics.\\[2pt]
Note: each rule is tagged [Type|Priority].\\
~~Types: P\,(Perceptual) / R\,(Reasoning) / C\,(Content) / F\,(Format) / L\,(Linguistic).\\
~~Priority: Critical / Important / Optional.\\[2pt]
[Rubrics]:\textcolor{blue}{\sout{<|im\_end|>}}
\par}
\end{tcolorbox}

\paragraph{Stage 3: Rule Quality Assessment.}
To ensure data fidelity, we prompt Seed2.0 Lite to score each rubric on a 1--5 scale along four dimensions:

\begin{itemize}[leftmargin=*,topsep=2pt,itemsep=3pt,parsep=0pt]
    \item \textbf{Specificity:} \textit{each rule is unambiguous and admits a clear binary judgment.}
    \item \textbf{Reasonableness:} \textit{each rule has genuine inspection value under the persona and task.}
    \item \textbf{Completeness:} \textit{the rule set covers the dimensions the scenario requires.}
    \item \textbf{Coherence:} \textit{the rules are mutually consistent and contradiction-free.}
\end{itemize}

A two-tier threshold then routes each rubric: a relaxed threshold admits it into the \emph{training pool}, while a strict threshold admits it into the \emph{evaluation pool} that seeds PRISM-Eval (\S\ref{sec:dataset}). Quality scoring uses the frozen Seed2.0 Lite model. For each $(I,\mathcal{R})$ it emits one JSON object with a 1--5 score and a short rationale per dimension. We parse the JSON and discard any rubric whose scoring response is malformed or incomplete, so a candidate enters the funnel below only if all four dimension scores are recovered.

Concretely, each rubric receives a score $s_d \in \{1,2,3,4,5\}$ for $d \in \{\text{Spec.}, \text{Reas.}, \text{Comp.}, \text{Coh.}\}$. The training pool admits a rubric iff
\[
\min_d s_d \ge 4 \quad \text{and} \quad \big| \{d : s_d = 5\} \big| \ge 2,
\]
while a strict threshold ($s_d = 5$ for all $d$) selects a candidate evaluation pool. We deduplicate this candidate pool against the training pool at both the image-source and rule-pattern level; the surviving rubrics form PRISM-Eval and are excluded from the training pool. This in-process gating shifts quality control from post-hoc curation to synthesis time, yielding a cleanly stratified rubric pool ready for downstream use.

\paragraph{Stage 4: Structured Response Generation.}
For each $(I, \mathcal{R})$ admitted to the training pool, we synthesize the supervision trace $\tau$ defined in \S\ref{sec:formulation} by prompting Seed2.0 Lite to emit a structured response with three explicit fields:

\begin{enumerate}[leftmargin=*,topsep=2pt,itemsep=3pt,parsep=0pt]
    \item \textbf{Visual Grounding ($g$):} A holistic description of $I$ that grounds the elements relevant to $\mathcal{R}$.
    \item \textbf{Rule-by-Rule Verification ($\{(j_i, e_i)\}$):} For each rule in \texttt{[Type|Priority]} order, the model analyzes the corresponding evidence in $I$ and emits a Pass/Fail judgment $j_i$ with a justification $e_i$.
    \item \textbf{Aggregation \& Conclusion ($y$):} A final verdict synthesized from per-rule judgments under priority precedence.
\end{enumerate}

The target overall verdict follows a fixed priority-aware policy. Any failed Critical rule is an immediate veto. If all Critical rules pass, Important and Optional rules receive weights $3$ and $1$, respectively:
\[
s=
\frac{3N_{\mathrm{Important,Pass}}+N_{\mathrm{Optional,Pass}}}
     {3N_{\mathrm{Important}}+N_{\mathrm{Optional}}},
\qquad
y=
\begin{cases}
\mathrm{Fail}, & \exists\, i:\ p_i=\mathrm{Critical}\ \land\ j_i=\mathrm{Fail},\\
\mathrm{Fail}, & s<0.5,\\
\mathrm{Pass}, & \text{otherwise}.
\end{cases}
\]
When no non-Critical rules are present, passing all Critical rules yields Pass. The formula is used to construct the supervision target but is intentionally not supplied at joint inference time: learning the mapping from prioritized rule judgments to the final verdict is part of the behavior PRISM is designed to internalize.

Rubric-based RL methods treat rubrics as a sparse reward signal over free-form outputs~\citep{gunjal2025rar,xu2026rubricarm,ruscarl2025}. We instead use rubrics as a \emph{structural template} for the learner's own reasoning trace, supplying fine-grained per-rule supervision. Type tags teach the learner to differentiate perceptual, inferential, content, format, and linguistic constraints. Priority annotations induce a verification ordering that prioritizes high-stakes rules at inference. The full output-format prompt is reproduced in Figure~\ref{fig:response-prompt}.

\subsection{Pipeline Instantiation}
\label{sec:dataset}

Applying the pipeline in \S\ref{sec:synthesis} to $\sim$250K Cambrian images---each treated as an independent sample yielding at most one rubric---produces $\sim$150K candidate rubrics, of which $\sim$110K pass Stage~3 quality filtering (a $72.6\%$ admission rate). These rubrics form two complementary pools that share the same generation procedure and structural schema but serve complementary roles.

\paragraph{PRISM-110K.}
PRISM-110K is the full supervision pool for fine-tuning, providing per-rule, type- and priority-aware training signals at scale. All downstream training subsets in \S\ref{sec:experiments}---both the $10$K used in main experiments and the $10$K--$80$K sweep used in the scaling study---are sampled uniformly at random from PRISM-110K without any rebalancing across types or priorities, so any observed performance bias reflects what the synthesis pipeline naturally produces rather than post-hoc curation.

\paragraph{PRISM-Eval.}
PRISM-Eval is a $1{,}000$-sample evaluation set used exclusively for assessment. It retains only the highest-quality rubrics (5/5 on all four dimensions) and is deduplicated against the training pool, so no evaluation rubric overlaps with training data. Scoring requires no inference-time judge: predictions are compared deterministically with fixed rule-level and overall labels under the protocol in \S\ref{sec:formulation}.

\paragraph{Yield statistics.}
Stage~2 produces approximately $150$K candidate rubrics; the remaining images are dropped because Stage~2 returns too few rules to constitute a usable rubric. Stage~3 then applies the two-tier quality filtering above, admitting roughly $72.6\%$ of the candidates. The $1{,}000$ rubrics that meet the strict threshold ($s_d{=}5$ for all $d$), after deduplication, form \textbf{PRISM-Eval}; the rubrics that meet only the relaxed threshold form \textbf{PRISM-110K}, our full supervision pool, from which all training subsets in \S\ref{sec:experiments} (e.g.\ the $10$K used in main experiments and the $10$K--$80$K used in the scaling study) are randomly sampled. PRISM-110K and PRISM-Eval are disjoint and share only the generation procedure, so there is no train--test overlap. The relaxed-vs-strict gap acts as a quality buffer: relaxed admission keeps the supervision pool abundant, while the strict cut reserves the cleanest rubrics for evaluation. Table~\ref{tab:funnel} summarizes this generation funnel.

\begin{figure}[H]
    \centering
    \includegraphics[width=\textwidth]{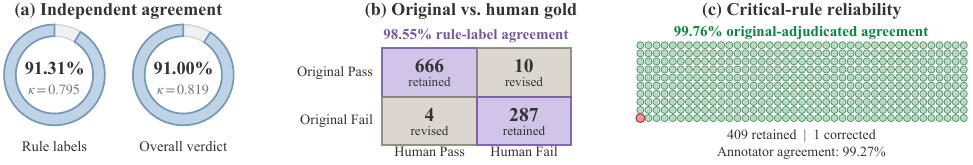}
    \caption{\textbf{Independent human validation of PRISM-Eval.}
    \textbf{(a)} Agreement between two independent annotators.
    \textbf{(b)} Original and adjudicated human rule labels agree on 98.55\% of 967 rules.
    \textbf{(c)} Critical rules show particularly high reliability: 409 of 410 original labels are retained.}
    \label{fig:human-audit}
    \vspace{-0.25cm}
\end{figure}

\paragraph{Independent human validation.}
\begin{wraptable}[10]{r}{0.5\textwidth}
\centering
\vskip -0.4cm
\small
\caption{PRISM data-generation funnel. Each Cambrian image is one independent sample yielding at most one rubric.}
\label{tab:funnel}
\begin{tabular*}{0.48\textwidth}{@{\extracolsep{\fill}}llr@{}}
\toprule
\textbf{Stage} & \textbf{Output} & \textbf{Count} \\
\midrule
Source images & Cambrian images & $\sim$250\,K \\
Stage 1--2 & Candidate rubrics & $\sim$150\,K \\
Stage 3 (relaxed) & PRISM-110K (train) & $\sim$110\,K \\
Stage 3 (strict) & PRISM-Eval (eval) & 1\,K \\
\bottomrule
\end{tabular*}
\vskip -0.3cm
\end{wraptable}
We validate a stratified 100-example subset of PRISM-Eval (967 rule labels). Two annotators relabel every rule and overall verdict blind to original gold labels; a third adjudicates disagreements. As Figure~\ref{fig:human-audit} shows, independent agreement reaches 91.31\% at rule level (Cohen's $\kappa{=}0.795$) and 91.00\% for overall verdicts ($\kappa{=}0.819$). Original labels agree with adjudicated human labels on 98.55\% of rules and 92.00\% of verdicts. Critical rules are especially stable: annotators agree on 99.27\%, and 409 of 410 original Critical labels are retained. The estimated correction rate is 1.45\% (95\% CI: 0.73--2.27\%).

Using the adjudicated labels as an alternative gold view yields the same conclusion as the full benchmark. On Qwen3-VL-4B, PRISM raises per-rule accuracy from 67.32\% to 82.11\%, Loose from 36.30\% to 70.90\%, and Strict from 8.70\% to 30.00\%. On Qwen3.5-9B, the corresponding scores rise from 70.37\% to 84.84\%, 49.70\% to 69.70\%, and 16.80\% to 33.40\%. Thus, the improvement persists when evaluation is anchored to independently adjudicated human labels.

\begin{figure*}[!t]
    \centering
    \begin{minipage}{.24\textwidth}
        \centering
        \includegraphics[width=.99\linewidth]{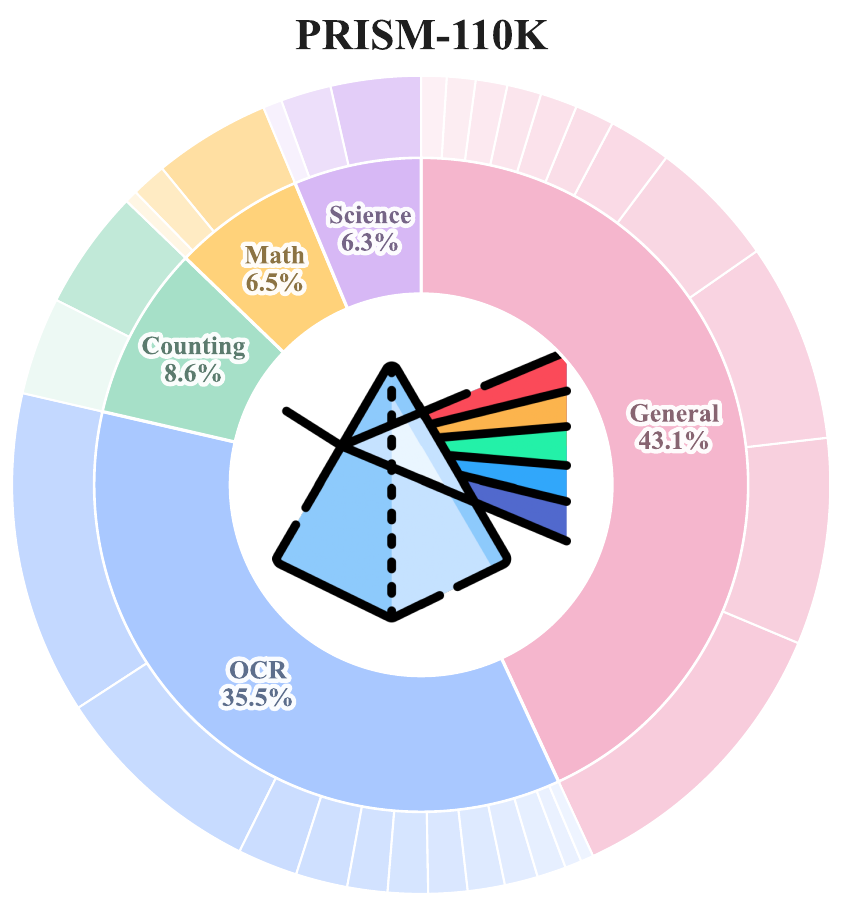}
    \end{minipage}%
    \hspace{0.002\textwidth}%
    \begin{minipage}{.755\textwidth}
        \centering
        \renewcommand{\arraystretch}{1.08}
        \fontsize{6pt}{9.3pt}\selectfont
        \newcommand{\dn}[1]{{\fontfamily{phv}\selectfont #1}}%
        \begin{minipage}[t]{0.29\linewidth}
        \vspace{0pt}
        \makecell{\cellcolor[RGB]{245,182,205} \textcolor{black}{\dn{\textbf{General}} \textbf{(47.4 K)}}}\\[2.5pt]
        \tikz[baseline=0.05em] \fill [color={rgb,255:red,248; green,198; blue,217}] (0,0) rectangle (0.75em,0.75em); \dn{ALLaVA}~\citep{chen2024allava} (12.3 K)\\
        \tikz[baseline=0.05em] \fill [color={rgb,255:red,249; green,204; blue,221}] (0,0) rectangle (0.75em,0.75em); \dn{COCO}~\citep{lin2014microsoft} (8.5 K)\\
        \tikz[baseline=0.05em] \fill [color={rgb,255:red,250; green,210; blue,225}] (0,0) rectangle (0.75em,0.75em); \dn{Q-Instruct}~\citep{wu2023q} (8.2 K)\\
        \tikz[baseline=0.05em] \fill [color={rgb,255:red,250; green,216; blue,229}] (0,0) rectangle (0.75em,0.75em); \dn{LNQA}~\citep{pont-tuset2019localizednarratives} (5.3 K)\\
        \tikz[baseline=0.05em] \fill [color={rgb,255:red,251; green,222; blue,232}] (0,0) rectangle (0.75em,0.75em); \dn{OODVQA}~\citep{tu2023many} (2.6 K)\\
        \tikz[baseline=0.05em] \fill [color={rgb,255:red,252; green,227; blue,235}] (0,0) rectangle (0.75em,0.75em); \dn{AlfWorld}~\citep{zhai2024fine} (1.6 K)\\
        \tikz[baseline=0.05em] \fill [color={rgb,255:red,252; green,232; blue,238}] (0,0) rectangle (0.75em,0.75em); \dn{VizWiz}~\citep{gurari2018vizwiz} (1.5 K)\\
        \tikz[baseline=0.05em] \fill [color={rgb,255:red,253; green,236; blue,240}] (0,0) rectangle (0.75em,0.75em); \dn{GPT4V-Rewriting} (1.4 K)\\
        \tikz[baseline=0.05em] \fill [color={rgb,255:red,253; green,240; blue,242}] (0,0) rectangle (0.75em,0.75em); \dn{IDK}~\citep{cha2024visually} (1.3 K)\\
        \tikz[baseline=0.05em] \fill [color={rgb,255:red,254; green,244; blue,245}] (0,0) rectangle (0.75em,0.75em); \dn{VisualGenome}~\citep{krishna2016visual} (1.2 K)\\
        \tikz[baseline=0.05em] \fill [color={rgb,255:red,254; green,247; blue,248}] (0,0) rectangle (0.75em,0.75em); \dn{GQA}~\citep{hudson2019gqa} (1.0 K)\\
        ~\\
        \end{minipage}
        \hspace{0.055\linewidth}
        \begin{minipage}[t]{0.29\linewidth}
        \vspace{0pt}
        \makecell{\cellcolor[RGB]{169,200,255} \textcolor{black}{\dn{\textbf{OCR}} \textbf{(39.1 K)}}}\\[2.5pt]
        \tikz[baseline=0.05em] \fill [color={rgb,255:red,179; green,206; blue,255}] (0,0) rectangle (0.75em,0.75em); \dn{DVQA}~\citep{kafle2018dvqa} (13.8 K)\\
        \tikz[baseline=0.05em] \fill [color={rgb,255:red,188; green,212; blue,255}] (0,0) rectangle (0.75em,0.75em); \dn{SynthDog}~\citep{kim2021donut} (9.2 K)\\
        \tikz[baseline=0.05em] \fill [color={rgb,255:red,197; green,219; blue,255}] (0,0) rectangle (0.75em,0.75em); \dn{ArxivQA}~\citep{li2024multimodal} (2.6 K)\\
        \tikz[baseline=0.05em] \fill [color={rgb,255:red,206; green,225; blue,255}] (0,0) rectangle (0.75em,0.75em); \dn{AI2D}~\citep{kembhavi2016diagram} (2.2 K)\\
        \tikz[baseline=0.05em] \fill [color={rgb,255:red,214; green,230; blue,255}] (0,0) rectangle (0.75em,0.75em); \dn{ScreenQA}~\citep{hsiao2022screenqa} (1.7 K)\\
        \tikz[baseline=0.05em] \fill [color={rgb,255:red,221; green,234; blue,255}] (0,0) rectangle (0.75em,0.75em); \dn{LLaVAR}~\citep{zhang2023llavar} (1.7 K)\\
        \tikz[baseline=0.05em] \fill [color={rgb,255:red,227; green,238; blue,255}] (0,0) rectangle (0.75em,0.75em); \dn{DocVQA}~\citep{mathew2021docvqa} (1.7 K)\\
        \tikz[baseline=0.05em] \fill [color={rgb,255:red,232; green,241; blue,255}] (0,0) rectangle (0.75em,0.75em); \dn{ChartQA}~\citep{masry2022chartqa} (1.6 K)\\
        \tikz[baseline=0.05em] \fill [color={rgb,255:red,237; green,244; blue,255}] (0,0) rectangle (0.75em,0.75em); \dn{WikiSQL}~\citep{zhong2017seq2sql} (1.4 K)\\
        \tikz[baseline=0.05em] \fill [color={rgb,255:red,242; green,247; blue,255}] (0,0) rectangle (0.75em,0.75em); \dn{OCRVQA}~\citep{mishra2019OCR} (1.2 K)\\
        \tikz[baseline=0.05em] \fill [color={rgb,255:red,254; green,247; blue,248}] (0,0) rectangle (0.75em,0.75em); \dn{WTQ}~\citep{pasupat2015compositional} (0.7 K)\\
        ~\\
        \end{minipage}
        \hspace{0.055\linewidth}
        \begin{minipage}[t]{0.29\linewidth}
        \vspace{0pt}
        \tikz[baseline=0.05em] \fill [color={rgb,255:red,255; green,250; blue,250}] (0,0) rectangle (0.75em,0.75em); \dn{IconQA}~\citep{lu2021iconqa} (0.5 K)\\
        \makecell{\cellcolor[RGB]{166,224,200} \textcolor{black}{\dn{\textbf{Counting}} \textbf{(9.5 K)}}}\\[2.5pt]
        \tikz[baseline=0.05em] \fill [color={rgb,255:red,178; green,231; blue,208}] (0,0) rectangle (0.75em,0.75em); \dn{CLEVR}~\citep{johnson2017clevr} (5.3 K)\\
        \tikz[baseline=0.05em] \fill [color={rgb,255:red,187; green,233; blue,212}] (0,0) rectangle (0.75em,0.75em); \dn{TallyQA}~\citep{acharya2019tallyqa} (4.3 K)\\
        \makecell{\cellcolor[RGB]{255,210,122} \textcolor{black}{\dn{\textbf{Math}} \textbf{(7.2 K)}}}\\[2.5pt]
        \tikz[baseline=0.05em] \fill [color={rgb,255:red,255; green,220; blue,143}] (0,0) rectangle (0.75em,0.75em); \dn{Geo170K}~\citep{gao2023g} (5.1 K)\\
        \tikz[baseline=0.05em] \fill [color={rgb,255:red,255; green,225; blue,155}] (0,0) rectangle (0.75em,0.75em); \dn{MathVision}~\citep{wang2024measuring} (1.5 K)\\
        \tikz[baseline=0.05em] \fill [color={rgb,255:red,255; green,230; blue,168}] (0,0) rectangle (0.75em,0.75em); \dn{RAVEN}~\citep{zhang2019raven} (0.5 K)\\
        \makecell{\cellcolor[RGB]{215,184,245} \textcolor{black}{\dn{\textbf{Science}} \textbf{(6.9 K)}}}\\[2.5pt]
        \tikz[baseline=0.05em] \fill [color={rgb,255:red,224; green,196; blue,248}] (0,0) rectangle (0.75em,0.75em); \dn{DataEngine} (3.8 K)\\
        \tikz[baseline=0.05em] \fill [color={rgb,255:red,228; green,202; blue,248}] (0,0) rectangle (0.75em,0.75em); \dn{PathVQA}~\citep{he2020pathvqa} (2.1 K)\\
        \tikz[baseline=0.05em] \fill [color={rgb,255:red,233; green,209; blue,249}] (0,0) rectangle (0.75em,0.75em); \dn{ScienceQA}~\citep{lu2022learn} (0.8 K)\\
        \end{minipage}
    \end{minipage}
    \caption{\textbf{PRISM-110K data composition.} \textbf{Left:} Domain-level distribution of PRISM-110K. The inner ring follows the verified domain proportions (General 43.1\%, OCR 35.5\%, Counting 8.6\%, Math 6.5\%, Science 6.3\%), while the outer ring preserves the corresponding child-dataset proportions within each domain. \textbf{Right:} All the data sources in the PRISM-110K as well as the ones filtered in data curation.}
    \vspace{-0.4cm}
    \label{fig:prism110k-data-mix}
\end{figure*}

\paragraph{Diversity \& composition.}
\begin{figure}[H]
    \centering
    \begin{minipage}[c]{0.49\textwidth}
        \centering
        \textbf{\footnotesize(a) Structural coverage}\\[3pt]
        \includegraphics[width=\linewidth]{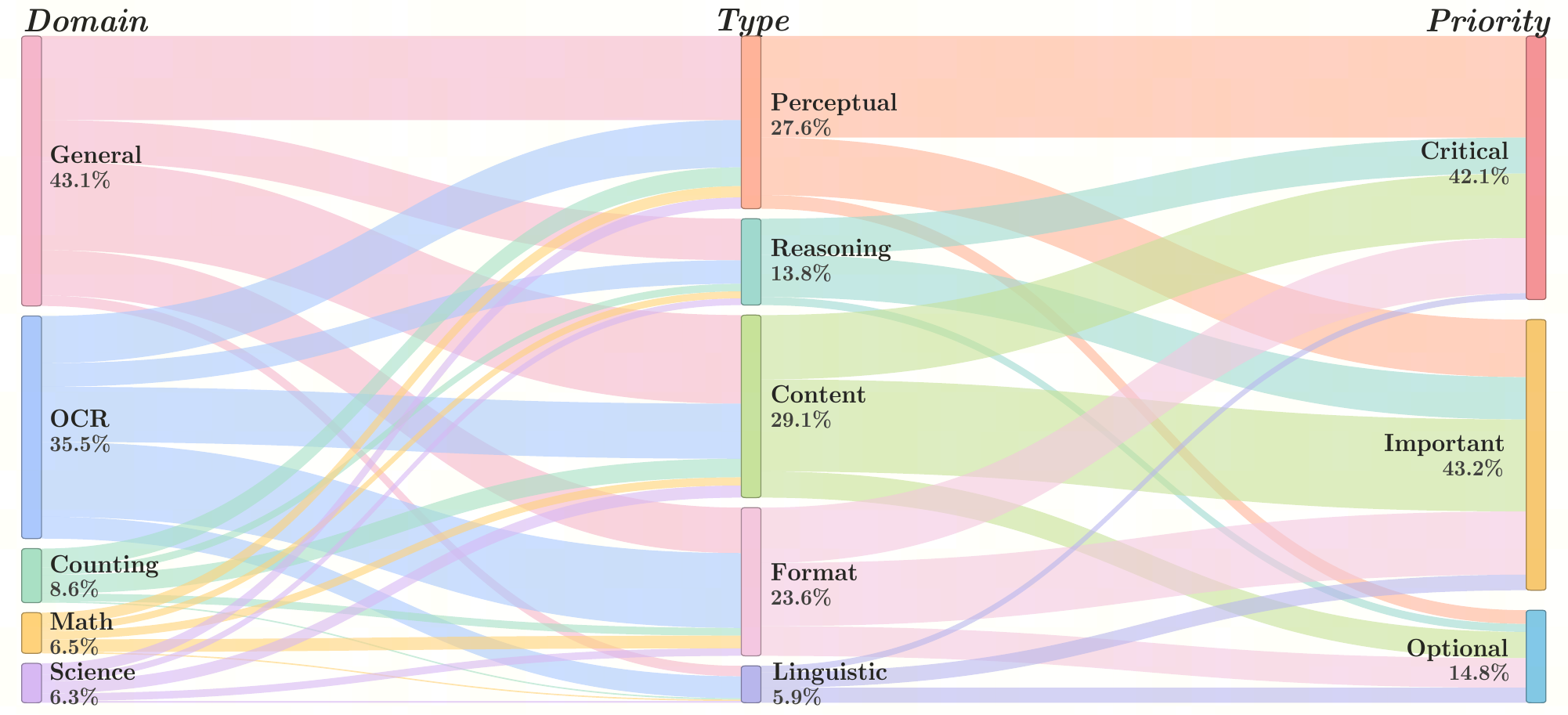}
    \end{minipage}\hfill
    \begin{minipage}[c]{0.49\textwidth}
        \centering
        \textbf{\footnotesize(b) Semantic diversity}\\[8pt]
        \footnotesize
        \setlength{\tabcolsep}{2.6pt}
        \renewcommand{\arraystretch}{1.20}
        \begin{tabular}{@{}lrrrr@{}}
        \toprule
        \textbf{Unit} & \textbf{$N$} & \textbf{Exact} &
        \multicolumn{2}{c}{\textbf{Max similarity}}\\
        \cmidrule(lr){4-5}
        & & & $\geq .95$ & $\geq .98$\\
        \midrule
        Task & 10K & 0.00 & 1.59 & $\approx$0.00\\
        Rule & 100{,}561 & 0.66 & 2.59 & 0.40\\
        Context.\ rule & 100{,}561 & 0.00 & 1.79 & 0.23\\
        \bottomrule
        \end{tabular}

    \end{minipage}
    \caption{\textbf{Diversity of PRISM-110K.}
    \textbf{(a)} The Sankey diagram connects image domain, rule type, and priority.
    \textbf{(b)} The adjacent table reports exact-duplicate and semantic near-neighbor rates (\%); contextualized rules include their persona--task context.}
    \label{fig:sankey}
    \vspace{-0.25cm}
\end{figure}
Figure~\ref{fig:sankey} examines diversity from complementary structural and semantic views. The pipeline covers all five rule types and three priority levels across General, OCR, Counting, Math, and Science images. Semantically, all 10,000 sampled persona--task pairs are unique; exact duplication is 0.66\% for standalone rules and 0\% after incorporating persona--task context. Only 2.59\% of rules have a cross-sample neighbor at cosine similarity at least 0.95, falling to 0.40\% at 0.98. Together, these results indicate that the synthesis process does not collapse to a small collection of repeated templates. Figure~\ref{fig:prism110k-data-mix} further summarizes the domain and source-dataset composition of PRISM-110K.

\paragraph{Engineering practices and efficiency.}
All four stages are stateless, per-image calls to frozen endpoints: Qwen3-VL-30B for Stage~2 and Seed2.0 Lite for Stages~1, 3, and 4. Because each image is processed independently, the pipeline is embarrassingly parallel: throughput scales near-linearly with the number of concurrent requests, and no cross-sample state or synchronization is required. With a request concurrency of $40$, generating the full corpus over $\sim$250K Cambrian images takes roughly $18$ hours end-to-end, including all four stages. Malformed or under-length responses are dropped rather than retried (the dominant Stage~2 failure mode is returning too few rules; see Table~\ref{tab:funnel}), which keeps the pipeline simple and avoids long-tail stalls. The same procedure can be applied to new image pools without retraining the synthesis models, although independent validation remains necessary when the source distribution changes.

\section{Experiments}
\label{sec:experiments}

\subsection{Experimental Setup}

\paragraph{Base Models.}
To assess whether PRISM's gains transfer across learner architectures, we run experiments on five MLLMs spanning the Qwen and InternVL families and a wide parameter range: Qwen3-VL-4B~\citep{baiQwen3VLTechnicalReport2025}, InternVL3.5-14B~\citep{wang2025internvl3}, and three Qwen3.5 variants---Qwen3.5-9B, Qwen3.5-27B, and Qwen3.5-35B-A3B~\citep{qwen3_5blog}---covering both dense and mixture-of-experts architectures from 4B to 35B.

\paragraph{Training Recipe.}
All experiments perform full-parameter SFT on top of LLaVA-NeXT-100K~\citep{li2024llavanext}, into which the per-method incremental data is mixed. To ensure a fair comparison, we adopt a single training recipe across every backbone and every comparison group: AdamW, peak learning rate $1\!\times\!10^{-5}$, cosine decay with $0.05$ warmup ratio, global batch size $128$, $1$ training epoch, and a maximum sequence length of $8192$. To improve compute efficiency on heterogeneous-length samples, we enable \emph{sequence packing}, which concatenates short samples up to the length budget so that no compute is wasted on padding tokens. We use full-parameter SFT throughout (no LoRA / no parameter freezing); the vision encoder is updated jointly with the language backbone.

\paragraph{Evaluation.}
We evaluate \textbf{in-domain} on \textbf{PRISM-Eval} and \textbf{out-of-domain} on 14 public benchmarks grouped into \textit{Perception}, \textit{Document}, and \textit{Reasoning} suites, whose unweighted average forms \textbf{Gen.~Avg}:
\begin{itemize}[leftmargin=1.2em,itemsep=2pt]
    \item \textbf{Perception:} MMBench\_EN~\citep{liu2024mmbench}, MME~\citep{fu2026mme}, MMStar~\citep{chen2024we}, MMMU~\citep{yue2024mmmu}, AI2D~\citep{kembhavi2016diagram}, covering broad visual perception, world knowledge, and diagram understanding.
    \item \textbf{Document:} OCRBench~\citep{liu2024ocrbench}, InfoVQA~\citep{mathew2022infographicvqa}, CharXiv descriptive and CharXiv reasoning~\citep{wang2024charxiv}, targeting OCR, text-rich images, and chart/document comprehension.
    \item \textbf{Reasoning:} DynaMath~\citep{zou2025dynamath}, MathVision\_MINI~\citep{wang2024measuring}, LogicVista~\citep{xiao2024logicvista}, VisuLogic~\citep{xu2025visulogic}, CountBenchQA~\citep{paiss2023countclip}, probing multimodal mathematical, logical, and counting reasoning.
\end{itemize}

The three suites are intentionally chosen to be \emph{disjoint} from rubric-style supervision, so that a non-trivial portion of each benchmark probes capabilities that PRISM never explicitly trains for, providing a single scalar that captures whether rubric-internalization training comes at the cost of general MLLM ability.

\paragraph{PRISM-Eval statistics and parsing.}
PRISM-Eval is nearly balanced at the sample level (505 Pass / 495 Fail overall verdicts) and contains 10,253 rule judgments (7,507 Pass / 2,746 Fail). An all-Pass predictor therefore obtains 50.5\% Loose but only 13.7\% Strict, illustrating why complete rule-level correctness is substantially more demanding than majority prediction. Our deterministic parser accepts semantically equivalent JSON object and list serializations, extracts rule judgments and the overall verdict separately, and counts any missing or invalid required label as incorrect. On Qwen3-VL-4B, parse failures occur for 27/1,000 Baseline outputs and 9/1,000 PRISM outputs; across the full leaderboard, the fraction ranges from 0.2\% for the best-formatted closed-source outputs to 7.0\% for LLaVA-family models.

\subsection{Leaderboard of PRISM-Eval}

\begin{wraptable}[20]{r}{0.52\textwidth}
\centering
\vskip -0.5cm
\footnotesize
\setlength{\tabcolsep}{2.4pt}
\renewcommand{\arraystretch}{0.90}
\caption{\textbf{Evaluation of various MLLMs on PRISM-Eval.} Loose/Strict accuracy (\%); section best in \textbf{bold}.}
\label{tab:leaderboard}
\begin{tabularx}{0.48\textwidth}{>{\raggedright\arraybackslash}X
                                  >{\centering\arraybackslash}p{0.07\textwidth}
                                  >{\centering\arraybackslash}p{0.065\textwidth}
                                  >{\centering\arraybackslash}p{0.065\textwidth}}
\toprule
\textbf{Model} & \multicolumn{1}{|c|}{\textbf{Param}} & \textbf{Loose} & \textbf{Strict} \\
\midrule
\multicolumn{4}{l}{\textbf{\it{Closed-Source MLLMs}}} \\
\midrule
\rowcolor{gray!10}
GPT-5.4~\citep{openaiGPT5} & \pcell{-} & \textbf{82.7} & \textbf{42.1} \\
\rowcolor{gray!10}
Gemini-3-Flash~\citep{gemini3flash} & \pcell{-} & 80.5 & 32.0 \\
\rowcolor{gray!10}
Seed2.0 Lite~\citep{seed2modelcard} & \pcell{-} & 77.8 & 38.3 \\
\midrule
\multicolumn{4}{l}{\textbf{\it{Open-Source MLLMs}}} \\
\midrule
\rowcolor{orange!12}
 & \pcell{2B} & 57.4 & 16.5 \\
\rowcolor{orange!12}
Qwen3-VL~\citep{baiQwen3VLTechnicalReport2025} & \pcell{4B} & 63.5 & 20.6 \\
\rowcolor{orange!12}
& \pcell{8B} & 67.1 & 23.5 \\
\rowcolor{red!10}
& \pcell{2B} & 57.0 & 15.9 \\
\rowcolor{red!10}
& \pcell{4B} & 59.0 & 21.5 \\
\rowcolor{red!10}
Qwen3.5~\citep{qwen3_5blog} & \pcell{9B} & 63.6 & 25.1 \\
\rowcolor{red!10}
 & \pcell{27B} & 63.9 & \textbf{29.3} \\
\rowcolor{red!10}
& \pcell{35B-A3B} & 61.8 & 25.1 \\
\rowcolor{blue!8}
& \pcell{2B} & 61.8 & 14.4 \\
\rowcolor{blue!8}
& \pcell{4B} & 64.0 & 15.9 \\
\rowcolor{blue!8}
InternVL3.5~\citep{wang2025internvl3} & \pcell{8B} & 62.8 & 16.8 \\
\rowcolor{blue!8}
 & \pcell{14B} & 63.6 & 18.3 \\
\rowcolor{blue!8}
& \pcell{30B-A3B} & \textbf{67.8} & 18.9 \\
\rowcolor{yellow!18}
LLaVA-1.5~\citep{li2024llava} & \pcell{7B} & 51.6 & 14.0 \\
\rowcolor{yellow!18}
LLaVA-NeXT~\citep{li2024llavanext} & \pcell{8B} & 52.3 & 14.3 \\
\bottomrule
\end{tabularx}
\vskip -0.2cm
\end{wraptable}

Before introducing PRISM training, we use PRISM-Eval to characterize the current state of rubric comprehension across representative closed- and open-source MLLMs. As reported in Table~\ref{tab:leaderboard}, we ask whether parameter scale alone can close the gap to closed-source frontiers.

Two findings stand out from Table~\ref{tab:leaderboard}. First, even GPT-5.4 reaches only $42.1\%$ Strict, and no open-source model crosses the $30\%$ Strict line. Second, larger backbones within the evaluated model families provide only modest Strict gains relative to the remaining gap. Thus, under the present protocol, model scale alone does not recover reliable per-rule, priority-aware judgment. These observations motivate a training signal that explicitly factorizes complex instructions into typed, prioritized rules and demands a verdict on each.

\begin{table*}[!t]
\centering
\caption{Main results. The upper block compares data synthesis methods on Qwen3-VL-4B (each adds $10$K on top of LLaVA-NeXT-100K); green deltas on the PRISM row are gains over Baseline. The lower block evaluates transfer of PRISM supervision across learner backbones (each row uses $10$K samples from PRISM-110K). Deltas are reported on Gen.~Avg and the in-domain PRISM-Eval metrics.}
\label{tab:main}
\resizebox{\textwidth}{!}{%
\begin{tabular}{l|c|cc|ccccc|ccc|ccccc}
\toprule
\multirow{2}{*}{\textbf{Model / Method}} & \multirow{2}{*}{\textbf{\makecell{General\\Avg}}} & \multicolumn{2}{c|}{\textbf{PRISM-Eval}} & \multicolumn{5}{c|}{\textbf{Perception}} & \multicolumn{3}{c|}{\textbf{Document}} & \multicolumn{5}{c}{\textbf{Reasoning}} \\
\cmidrule(lr){3-4} \cmidrule(lr){5-9} \cmidrule(lr){10-12} \cmidrule(lr){13-17}
& & Loose & Strict & MMB & MME & MMS & MMMU & AI2D & OCR & Info & ChXiv$_{d/r}$ & Dyn & MV & Logic & Visu & Count \\
\midrule
\rowcolor{gray!12}
\multicolumn{17}{l}{\textit{Comparison with data synthesis methods (Qwen3-VL-4B, LLaVA-NeXT-100K $+10$K synthesized samples)}} \\
\midrule
Baseline & 60.30 & 36.7 & 9.5 & 84.7 & 83.0 & 62.7 & 53.7 & 84.4 & 79.5 & 69.8 & 70.6/38.6 & 41.4 & 23.0 & 39.8 & 24.3 & 88.7 \\
LLaVA-CoT & 60.68 & 53.0 & 14.7 & 84.0 & 83.0 & 62.1 & 55.0 & 84.8 & 80.9 & 70.4 & 71.9/39.0 & 42.2 & 22.4 & 38.0 & 25.5 & 90.3 \\
MMEvol & 60.66 & 51.4 & 14.2 & 83.6 & 83.6 & 62.2 & 54.9 & 84.0 & 81.3 & 69.9 & 73.2/38.0 & 41.2 & 22.0 & 39.6 & 26.7 & 89.1 \\
Mulberry & 60.69 & 54.7 & 14.8 & 83.4 & 83.6 & 63.0 & 55.1 & 84.4 & 81.4 & 69.6 & 71.4/37.6 & 41.4 & 22.0 & 39.1 & 27.4 & 90.3 \\
Cambrian & 60.34 & 52.9 & 14.9 & 83.5 & 83.1 & 61.7 & 52.3 & 83.5 & 82.3 & 70.5 & 73.6/37.9 & 41.7 & 24.7 & 37.8 & 23.9 & 88.9 \\
OASIS & 60.57 & 54.5 & 15.0 & 84.5 & 83.5 & 63.4 & 54.0 & 84.4 & 80.5 & 69.3 & 71.2/38.6 & 41.1 & 22.4 & 38.3 & 25.6 & 90.1 \\
\rowcolor{blue!6}
\textbf{PRISM (Ours)} & 60.87 \textcolor{ForestGreen}{\scriptsize+0.6} & 71.1 \textcolor{ForestGreen}{\scriptsize+34.4} & 30.1 \textcolor{ForestGreen}{\scriptsize+20.6} & 85.0 & 83.0 & 62.5 & 53.8 & 83.9 & 80.5 & 68.8 & 72.6/39.7 & 41.6 & 24.7 & 39.4 & 25.9 & 90.8 \\
\midrule
\rowcolor{gray!12}
\multicolumn{17}{l}{\textit{Transfer across learner backbones (LLaVA-NeXT-100K $+10$K synthesized samples)}} \\
\midrule
InternVL3.5-14B & 65.10 & 65.8 & 19.7 & 84.9 & 86.3 & 67.2 & 61.9 & 85.6 & 82.4 & 77.9 & 80.5/43.6 & 49.7 & 32.2 & 46.8 & 27.1 & 85.4 \\
\rowcolor{blue!6}
\quad \textbf{w.\ PRISM} & 65.56 \textcolor{ForestGreen}{\scriptsize+0.5} & 66.3 \textcolor{ForestGreen}{\scriptsize+0.5} & 28.8 \textcolor{ForestGreen}{\scriptsize+9.1} & 85.2 & 85.5 & 67.9 & 61.8 & 85.8 & 82.4 & 78.8 & 81.2/44.1 & 49.4 & 34.5 & 49.9 & 27.5 & 83.8 \\
\midrule
Qwen3.5-9B & 66.63 & 49.5 & 16.6 & 87.5 & 87.4 & 68.1 & 57.8 & 89.1 & 87.5 & 73.8 & 79.0/57.4 & 61.2 & 27.3 & 43.8 & 24.8 & 88.1 \\
\rowcolor{blue!6}
\quad \textbf{w.\ PRISM} & 68.17 \textcolor{ForestGreen}{\scriptsize+1.5} & 69.6 \textcolor{ForestGreen}{\scriptsize+20.1} & 33.2 \textcolor{ForestGreen}{\scriptsize+16.6} & 87.4 & 89.0 & 69.9 & 60.3 & 89.1 & 87.2 & 74.0 & 80.0/57.2 & 71.8 & 27.6 & 44.7 & 27.5 & 88.7 \\
\midrule
Qwen3.5-35B-A3B & 68.83 & 62.8 & 23.9 & 88.6 & 88.6 & 69.3 & 62.6 & 90.8 & 87.8 & 78.3 & 81.4/59.6 & 68.3 & 27.6 & 43.8 & 27.4 & 89.5 \\
\rowcolor{blue!6}
\quad \textbf{w.\ PRISM} & 70.45 \textcolor{ForestGreen}{\scriptsize+1.6} & 68.9 \textcolor{ForestGreen}{\scriptsize+6.1} & 36.2 \textcolor{ForestGreen}{\scriptsize+12.3} & 88.7 & 89.1 & 71.1 & 64.3 & 90.4 & 88.4 & 79.4 & 83.6/60.8 & 77.6 & 29.6 & 44.7 & 28.5 & 90.1 \\
\midrule
Qwen3.5-27B & 71.61 & 63.6 & 27.8 & 88.9 & 89.1 & 73.7 & 68.2 & 91.2 & 87.5 & 80.1 & 83.0/61.1 & 72.2 & 36.5 & 52.6 & 26.8 & 91.6 \\
\rowcolor{blue!6}
\quad \textbf{w.\ PRISM} & 72.52 \textcolor{ForestGreen}{\scriptsize+0.9} & 72.3 \textcolor{ForestGreen}{\scriptsize+8.7} & 38.4 \textcolor{ForestGreen}{\scriptsize+10.6} & 89.8 & 91.1 & 72.5 & 66.6 & 91.0 & 88.4 & 80.1 & 85.9/63.2 & 79.9 & 32.6 & 52.8 & 27.6 & 93.8 \\
\bottomrule
\end{tabular}%
}
\vspace{-0.2cm}
\end{table*}

\subsection{Main Results}

\paragraph{Comparison with Data Synthesis Baselines.}
We first ask whether existing data-synthesis recipes can already impart rubric comprehension. The upper block of Table~\ref{tab:main} compares PRISM with five representative baselines on Qwen3-VL-4B, all sharing the same recipe and incremental data volume atop LLaVA-NeXT-100K: \textbf{LLaVA-CoT}~\citep{xu2024llavacot} (visual instruction data emphasizing reasoning chains), \textbf{MMEvol}~\citep{luo2024mmevol} (iteratively evolved instruction data), \textbf{Mulberry}~\citep{yao2024mulberry} (collective reasoning data emphasizing path diversity), \textbf{Cambrian}~\citep{tong2024cambrian} (curated visual QA), and \textbf{OASIS}~\citep{zhang2025oasis} (prefix completion without prompt templates). All five cluster tightly around the no-incremental-data baseline on PRISM-Eval. Increasing the volume of generic visual instruction data, however diversified, does not by itself induce per-rule, priority-aware judgment. PRISM departs from this cluster because it injects \emph{structural} supervision (typed rules, priority, and per-rule verification) rather than additional samples. Crucially, this in-domain gain leaves Gen.~Avg essentially unchanged, indicating no measurable degradation on general benchmarks. Finer-grained breakdowns by priority, rule type, and rubric size are provided in \S\ref{sec:error-breakdown}, while controlled format variants in \S\ref{sec:format-robustness} test whether the gain persists under changes to the rubric representation.

\paragraph{Transfer across Learner Backbones.}
The lower block tests whether this effect is backbone-specific. Adding the same $10$K synthesized samples to four MLLMs spanning 9B--35B parameters in both dense and MoE configurations raises Strict accuracy on \emph{every} backbone, while leaving Gen.~Avg unchanged or slightly higher. Since PRISM contributes only a tiny fraction of the total training tokens, this directional consistency suggests that rubric comprehension occupies a capability axis largely orthogonal to what these backbones absorb during pre-training---structured supervision can be slotted in without disturbing existing skills. Notably, the strongest configuration rises well above the best open-source result in Table~\ref{tab:leaderboard} and narrows the gap to the closed-source models with only 10K synthesized samples, suggesting that the binding constraint is the \emph{kind} of supervision, not the parameter count.

\FloatBarrier

\subsection{What Makes PRISM Work?}
\label{sec:ablation}

We isolate each design choice in PRISM along three axes---\emph{data} (what enters the model), \emph{procedure} (how the model reasons), and \emph{rubric structure} (what the labels carry). All variants are trained with the same recipe on Qwen3-VL-4B with $10$K synthesized samples; results are summarized in Table~\ref{tab:ablation}.

\paragraph{Data vs.\ Procedure.}
Removing \textbf{quality filtering} barely affects Loose accuracy but lowers Strict and Gen.~Avg, consistent with \emph{data-purity} interference: low-quality rubrics rarely flip the overall verdict (so Loose is robust), but they inject ambiguous or self-contradictory rules that degrade per-rule judgment. Removing \textbf{structured CoT} shows the opposite asymmetry: Loose and Gen.~Avg fall the most, while Strict drops by the same margin as the filtering ablation. This is consistent with structured CoT acting as a \emph{procedural prior} (Visual Grounding $\rightarrow$ Rule-by-Rule $\rightarrow$ Aggregation) that keeps reasoning aligned with the task structure across both rubric and non-rubric inputs; without it, the model still recognizes individual rules at strict granularity but loses the holistic verification habit that general-benchmark transfer relies on. The two effects are complementary rather than redundant---filtering safeguards what enters the model, structured CoT shapes how the model reasons over what it has learned.

\paragraph{Priority vs.\ Order.}
The remaining two rows probe what the rubric structure itself contributes. Stripping the \textbf{Priority} field while keeping types and the per-rule trace causes the largest single-component drop in our entire ablation: Strict falls by $-6.5$\,pp and Loose by $-4.3$\,pp, far exceeding either of the data- or procedure-side ablations. Without priority annotations, the model treats every rule as equally decisive and loses the Critical-first verification ordering induced at Stage~4 (\S\ref{sec:synthesis}); the resulting drop is precisely the gain we will dissect by Priority bucket in \S\ref{sec:error-breakdown}. In contrast, when we retrain on rubrics whose rules have been \textbf{randomly shuffled} within each sample (types and priorities preserved), performance is statistically indistinguishable from PRISM Full ($30.3$ vs.\ $30.1$ Strict, $70.5$ vs.\ $71.1$ Loose). The two rows draw a clean line: it is the \emph{semantic} structure of priority labels that PRISM depends on, not the \emph{surface} order in which rules happen to be listed.

\begin{table*}[!t]
\centering
\caption{Component ablations on Qwen3-VL-4B ($10$K synthesized samples). Red deltas on General Avg and PRISM-Eval mark drops relative to PRISM (Full); the Shuffled-rules row controls for surface ordering rather than removing a component.}
\label{tab:ablation}
\resizebox{\textwidth}{!}{%
\begin{tabular}{l|c|cc|ccccc|ccc|ccccc}
\toprule
\multirow{2}{*}{\textbf{Configuration}} & \multirow{2}{*}{\textbf{\makecell{General\\Avg}}} & \multicolumn{2}{c|}{\textbf{PRISM-Eval}} & \multicolumn{5}{c|}{\textbf{Perception}} & \multicolumn{3}{c|}{\textbf{Document}} & \multicolumn{5}{c}{\textbf{Reasoning}} \\
\cmidrule(lr){3-4} \cmidrule(lr){5-9} \cmidrule(lr){10-12} \cmidrule(lr){13-17}
& & Loose & Strict & MMB & MME & MMS & MMMU & AI2D & OCR & Info & ChXiv$_{d/r}$ & Dyn & MV & Logic & Visu & Count \\
\midrule
\rowcolor{ForestGreen!8}
\textbf{PRISM (Full)} & \textbf{60.87} & \textbf{71.1} & \textbf{30.1} & 85.0 & 83.0 & 62.5 & 53.8 & 83.9 & 80.5 & 68.8 & 72.6/39.7 & 41.6 & 24.7 & 39.4 & 25.9 & 90.8 \\
\midrule
\multicolumn{17}{l}{\textit{Data side: what enters the model.}} \\
\quad w/o Quality Filtering
& 60.49\,\textcolor{red}{\scriptsize-0.38}
& 71.0\,\textcolor{red}{\scriptsize-0.1}
& 29.4\,\textcolor{red}{\scriptsize-0.7}
& 84.5 & 82.5 & 62.7 & 53.3 & 83.8 & 81.8 & 69.2 & 71.9/38.1 & 40.9 & 24.0 & 39.3 & 25.4 & 89.4 \\
\midrule
\multicolumn{17}{l}{\textit{Procedure side: how the model reasons.}} \\
\quad w/o Structured CoT
& 60.10\,\textcolor{red}{\scriptsize-0.77}
& 69.2\,\textcolor{red}{\scriptsize-1.9}
& 29.4\,\textcolor{red}{\scriptsize-0.7}
& 84.5 & 82.4 & 62.5 & 52.7 & 84.2 & 79.9 & 69.7 & 71.1/38.5 & 40.9 & 21.7 & 39.2 & 24.8 & 89.3 \\
\midrule
\multicolumn{17}{l}{\textit{Rubric structure: semantic priority vs.\ surface order.}} \\
\quad w/o Priority
& 60.45\,\textcolor{red}{\scriptsize-0.42}
& 66.8\,\textcolor{red}{\scriptsize-4.3}
& 23.6\,\textcolor{red}{\scriptsize-6.5}
& 85.3 & 85.8 & 62.4 & 53.9 & 84.2 & 81.5 & 68.0 & 71.2/39.1 & 41.0 & 21.4 & 40.3 & 25.1 & 89.9 \\
\quad Shuffled rules
& 60.84\,\textcolor{red}{\scriptsize-0.03}
& 70.5\,\textcolor{red}{\scriptsize-0.6}
& 30.3\,\textcolor{ForestGreen}{\scriptsize+0.2}
& 84.7 & 84.6 & 62.1 & 54.1 & 84.4 & 81.3 & 68.7 & 72.2/38.5 & 41.9 & 22.7 & 41.6 & 26.3 & 91.0 \\
\bottomrule
\end{tabular}%
}
\vspace{-0.3cm}
\end{table*}

\paragraph{Aggregation Internalization.}
\begin{wraptable}[9]{r}{0.40\textwidth}
\centering
\vskip -0.5cm
\caption{\textbf{Aggregation diagnostic on Qwen3-VL-4B} (\%). Generated/re-aggregated verdict accuracy and generated-verdict accuracy when all rule predictions are correct.}
\label{tab:aggregation-diagnostic}
\fontsize{8.0pt}{9.6pt}\selectfont
\setlength{\tabcolsep}{2.6pt}
\renewcommand{\arraystretch}{1.18}
\begin{tabular*}{0.38\textwidth}{@{\extracolsep{\fill}}lcc@{}}
\toprule
\textbf{Model} & \textbf{\makecell{Generated /\\Re-aggregated}} & \textbf{\makecell{All rules\\correct}} \\
\midrule
Baseline & 36.7 \resultarrow 47.5\;\textcolor{ForestGreen}{\scriptsize +10.8} & 66.4 \\
PRISM & \textbf{71.1} \resultarrow \textbf{73.6}\;\textcolor{ForestGreen}{\scriptsize +2.5} & \textbf{87.8} \\
\bottomrule
\end{tabular*}
\vskip -0.2cm
\end{wraptable}
The aggregation formula is deliberately withheld at inference time because mapping prioritized rule judgments to the final verdict is part of the capability that PRISM is intended to learn. To isolate this component, we perform an oracle-policy diagnostic: we retain each model's predicted rule labels, discard its generated overall verdict, and recompute that verdict using the fixed policy in \S\ref{sec:synthesis}. This diagnostic gives the policy to an external program and is therefore not an equal-condition baseline. As Table~\ref{tab:aggregation-diagnostic} shows, programmatic re-aggregation raises the Baseline's Loose accuracy from 36.7\% to 47.5\% ($+10.8$\,pp), but changes PRISM only from 71.1\% to 73.6\% ($+2.5$\,pp). When every predicted rule label is correct, PRISM's own overall verdict is correct in 302/344 cases (87.8\%), compared with 95/143 (66.4\%) for the Baseline. The smaller oracle-policy gain and higher conditional consistency indicate that PRISM has largely internalized the priority-aware aggregation behavior.

\FloatBarrier

\subsection{Where Do the Gains Come From?}
\label{sec:error-breakdown}

The headline results in Table~\ref{tab:main} report a single Strict number per model, leaving open \emph{where} the gains come from. We dissect PRISM-Eval errors along three orthogonal axes---rubric \emph{Priority}, rule \emph{Type}, and rubric \emph{Size} (number of rules). Because the same evaluation set carries information at three different granularities, we report per-rule accuracy for Priority and Type, which operate on rule instances, and Strict accuracy for Size, which operates on complete samples.

We compare the Qwen3-VL-4B \textbf{Baseline} (LLaVA-NeXT-100K only), our \textbf{PRISM}-trained variant (Baseline $+\,10$K PRISM samples; the same checkpoint reported in Table~\ref{tab:main}), and the strongest closed-source model on the leaderboard, \textbf{GPT-5.4}.

\begin{figure}[H]
    \centering
    \includegraphics[width=0.95\textwidth]{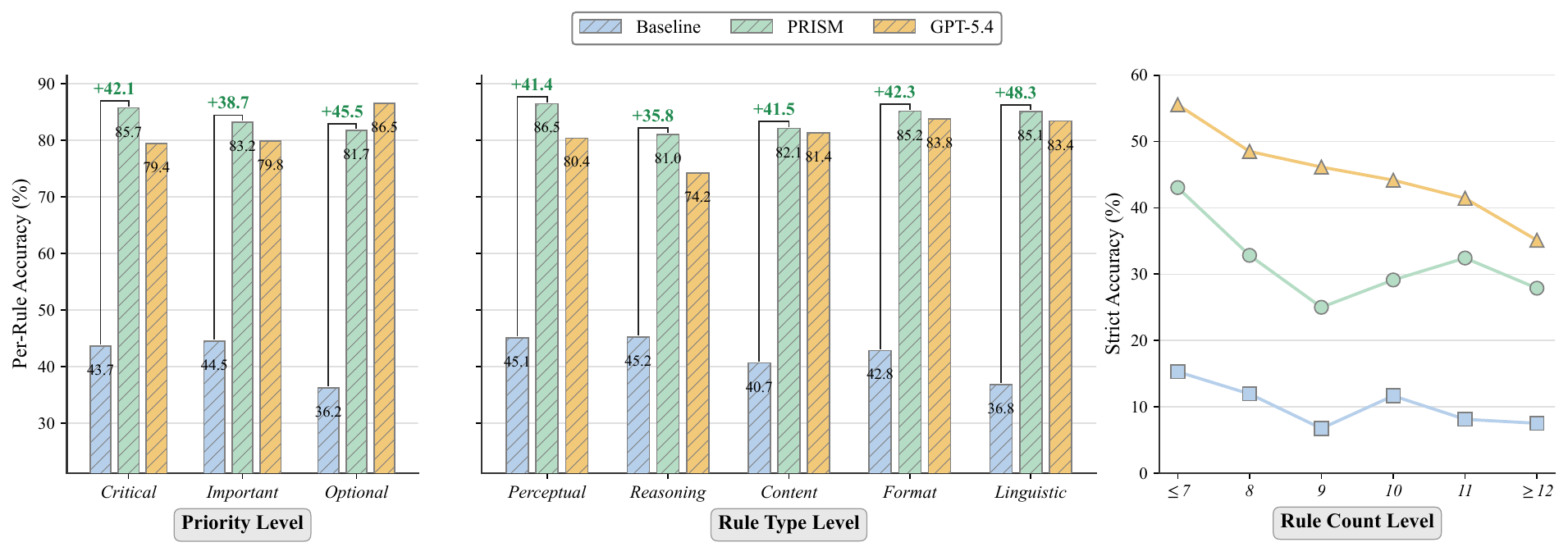}
    \caption{\textbf{Error breakdown on PRISM-Eval.} \textbf{Baseline} denotes Qwen3-VL-4B fine-tuned on LLaVA-NeXT-100K, and \textbf{PRISM} denotes the same backbone fine-tuned on LLaVA-NeXT-100K $+\,10$K PRISM samples; \textbf{GPT-5.4} is the strongest closed-source reference. We report per-rule accuracy stratified by \emph{Priority Level} (left), per-rule accuracy stratified by \emph{Rule Type Level} (middle), and per-sample Strict accuracy stratified by \emph{Rule Count Level} (right).}
    \vspace{-0.3cm}
    \label{fig:error-breakdown}
\end{figure}

\paragraph{By Priority.}
PRISM lifts per-rule accuracy across all three priority buckets: from 43.7\% to 85.7\% on Critical, 44.5\% to 83.2\% on Important, and 36.2\% to 81.7\% on Optional. The resulting gains of $+42.1$, $+38.7$, and $+45.5$\,pp show that the improvement is broad rather than confined to one priority level. Critical rules nevertheless attain the highest absolute accuracy, which matters because any Critical failure can determine the overall verdict. Together with the $-6.5$\,pp Strict drop in the \textbf{w/o Priority} ablation, these results support the value of explicitly encoding priority semantics.

\paragraph{By Rule Type.}
The Baseline's weakest axes are \emph{Reasoning} and \emph{Content}---both require composing visual evidence into a fact-grounded judgment beyond mere perception. PRISM substantially reduces per-rule error on both, and also improves Format and Linguistic rules. This supports the role of the typed taxonomy in \S\ref{sec:synthesis}: explicit \texttt{[Type]} tags push the model to differentiate Perceptual ("check what is in the image"), Reasoning ("infer over it"), and Content ("check topic coverage"), rather than collapsing them into a single answering mode. GPT-5.4 also lags PRISM on Reasoning in this evaluation, suggesting the axis remains challenging and that typed supervision directly targets it.

\paragraph{By Rule Count.}
Strict accuracy declines mechanically with rubric size, since every rule must be judged correctly. What matters is the shape of the gap: PRISM beats the Baseline by a wide margin on small rubrics ($\leq 7$), and the lead persists at $\geq 12$, where the Baseline approaches zero. The Strict gain in Table~\ref{tab:main} is therefore not an artifact of short rubrics; the advantage remains visible as the number of jointly evaluated rules increases.

\FloatBarrier

\subsection{How Far Does Structured Supervision Scale?}
\label{sec:scaling}

\begin{wrapfigure}[10]{r}{0.5\textwidth}
    \centering
    \vskip -0.5cm
    \includegraphics[width=0.48\textwidth]{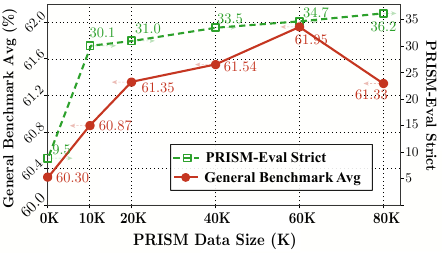}
    \setlength{\abovecaptionskip}{-2pt}   
    \caption{Scaling curves of PRISM on Qwen3-VL-4B.}
    \label{fig:scaling}
    \vskip -0.5cm
\end{wrapfigure}
We investigate the effect of training-set size (from $10$K to $80$K synthesized samples) on Qwen3-VL-4B performance. The two curves diverge in shape. \textbf{Gen.~Avg follows a shallow inverted-U}, peaking at a moderate scale before retreating, whereas \textbf{in-domain Strict increases monotonically}. This pattern is consistent with the general-benchmark benefit saturating after the structured ``ground--verify--aggregate'' procedure is learned, while additional rubric data continues to improve fidelity on the target task. In practice, a moderate pool gives the best Gen.~Avg trade-off, whereas larger pools favor in-domain Strict accuracy.


\FloatBarrier

\subsection{How Robust Is PRISM to Equivalent Representations?}
\label{sec:format-robustness}

\begin{wraptable}[8]{r}{0.49\textwidth}
\centering
\vskip -0.45cm
\caption{\textbf{Strict accuracy under controlled representation changes} (\%). Base \(\rightarrow\) PRISM.}
\label{tab:format-robustness}
\fontsize{8.0pt}{9.6pt}\selectfont
\setlength{\tabcolsep}{2.4pt}
\renewcommand{\arraystretch}{1.15}
\begin{tabular*}{0.47\textwidth}{@{\extracolsep{\fill}}lcc@{}}
\toprule
\textbf{Variant} & \textbf{4B} & \textbf{9B} \\
\midrule
Original & 9.5 \resultarrow \textbf{30.1}\;\textcolor{ForestGreen}{\scriptsize +20.6} & 16.6 \resultarrow \textbf{33.2}\;\textcolor{ForestGreen}{\scriptsize +16.6} \\
Paraphrase & 8.4 \resultarrow \textbf{30.2}\;\textcolor{ForestGreen}{\scriptsize +21.8} & 4.2 \resultarrow \textbf{35.5}\;\textcolor{ForestGreen}{\scriptsize +31.3} \\
Nat.-lang. tags & 3.0 \resultarrow \textbf{30.4}\;\textcolor{ForestGreen}{\scriptsize +27.4} & 5.2 \resultarrow \textbf{29.0}\;\textcolor{ForestGreen}{\scriptsize +23.8} \\
Output format & 10.0 \resultarrow \textbf{17.2}\;\textcolor{ForestGreen}{\scriptsize +7.2} & 20.7 \resultarrow \textbf{23.7}\;\textcolor{ForestGreen}{\scriptsize +3.0} \\
\bottomrule
\end{tabular*}
\vskip -0.2cm
\end{wraptable}
We construct three controlled PRISM-Eval variants, changing exactly one factor at a time while preserving images, rule identities, priorities, and gold judgments: \textbf{Paraphrase} rewrites each rule with equivalent wording; \textbf{Natural-language tags} replaces compact \texttt{[Type|Priority]} markers with explicit natural-language descriptions; and \textbf{Output format} replaces the required JSON serialization with a line-based rule list. Table~\ref{tab:format-robustness} compares each PRISM-trained model with its corresponding Baseline. PRISM retains a positive Strict gain in every condition on both 4B and 9B. The gains are particularly strong under rule paraphrasing and natural-language tags, showing that the improvement is not tied to fixed rule wording or compact tag tokens. Transfer to a new output schema is positive but more modest and model-dependent, separating input-side rubric comprehension from output-serialization difficulty.

\FloatBarrier

\subsection{Can Inference-Time Strategies Replace Training?}
\label{sec:training-free}

\begin{figure}[H]
    \centering
    \includegraphics[width=0.92\textwidth]{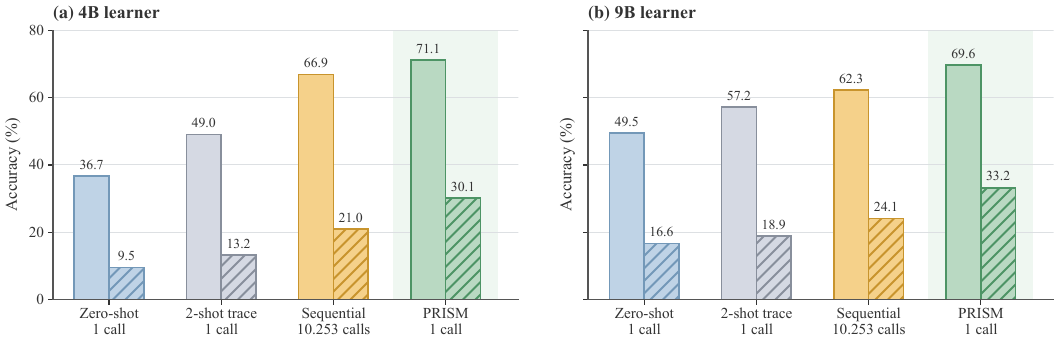}
    \caption{\textbf{Training-free inference strategies on PRISM-Eval.}
    Two-shot trace prompting and Sequential rule-wise querying recover part of the capability, but PRISM achieves the highest Loose and Strict accuracy with a single joint call. Hatching denotes Strict accuracy.}
    \label{fig:training-free}
    \vspace{-0.2cm}
\end{figure}

We evaluate two training-free alternatives on the same 4B and 9B Baseline checkpoints. \textbf{2-shot trace prompting} prepends one Pass and one Fail training exemplar, each demonstrating visual grounding, rule-wise verification, and aggregation. The stronger \textbf{Sequential} strategy queries every rule independently and applies the explicit priority-aware policy programmatically. As Figure~\ref{fig:training-free} shows, both approaches improve over joint zero-shot inference, confirming that demonstrations and decomposition are useful. Nevertheless, PRISM remains substantially stronger in Strict accuracy: 30.1\% vs.\ 21.0\% for Sequential on 4B, and 33.2\% vs.\ 24.1\% on 9B. Sequential also requires one call per rule---10.253 calls per sample on average---plus external aggregation, whereas PRISM jointly predicts all rule judgments and the overall verdict in one call without receiving the aggregation formula at inference time. Thus, inference-time structure recovers part of the capability, but does not match the accuracy or efficiency of internalizing the procedure through training.

\FloatBarrier

\section{Conclusion}

We formulate \textbf{rubric comprehension} as an executor-side task that requires a model to verify multiple prioritized rules and produce an overall judgment within a single multimodal instruction. We introduce \textbf{PRISM}, a four-stage data synthesis framework that generates typed, priority-aware rubrics and structured verification traces, yielding the training pool \textbf{PRISM-110K} and the instance-disjoint evaluation set \textbf{PRISM-Eval}. Across five MLLM backbones spanning dense and MoE architectures from 4B to 35B, PRISM improves in-domain Strict accuracy by 9.1--20.6 percentage points without measurable degradation on the evaluated general benchmarks. Independent human adjudication supports the reliability of the fixed labels, controlled representation variants demonstrate robustness beyond the original rule wording and tags, and training-free baselines recover only part of the gain at greater inference cost. Together, these results establish structured rubric supervision as an effective and scalable approach to priority-aware multimodal verification.

\section*{Limitations}

PRISM currently relies on supervised fine-tuning, leaving reinforcement learning or preference optimization with rule-level feedback unexplored. Our experiments focus on single-image, static verification; extending structured rubric supervision to video and multi-turn settings will require temporal and interaction-aware rubric formulations. Finally, exact all-rule correctness remains challenging as rubric size increases, motivating stronger visual reasoning and more reliable tracking of multiple constraints.

\bibliographystyle{unsrtnat}
\bibliography{custom,reference}

\clearpage
\appendix

\begin{figure*}[p]
\centering
\vspace*{\fill}
\begin{tcolorbox}[
    colback=RoyalBlue!4,
    colframe=RoyalBlue!75!black,
    coltitle=white,
    fonttitle=\bfseries,
    title={Prompt Used for Persona--Task Generation},
    boxrule=0.6pt,
    arc=2pt,
    left=6pt, right=6pt, top=4pt, bottom=4pt,
    before skip=0pt, after skip=0pt
]
\textbf{[SYSTEM PROMPT]}\\[2pt]
{\fontsize{8pt}{10pt}\selectfont\ttfamily
You are a multimodal task design expert. Your task is to design a reasonable user persona and a complex rule-based judgment task for a given image.\\[4pt]
\#\# Core Concept\\[2pt]
A complex rule-based judgment task refers to: given an image, requiring a systematic inspection of the image content based on multiple rules, and providing a comprehensive judgment result. Such tasks are widely present in real-world scenarios like content moderation, compliance checking, and quality assessment.\\[4pt]
\#\# Generation Strategy\\[2pt]
Please reason step-by-step following this chain of thought:\\[4pt]
\textbf{Step 1: Analyze Image Intent.} First, understand the ``usage intent'' of this image---that is, the functional purpose for which it was created or used. Image intent is not a description of visual content (e.g., ``a blue poster'') or physical category (e.g., ``indoor photo''); rather, it addresses: why was this image produced, what purpose might it serve in a real-world scenario, and who are its likely target audience or users?\\[2pt]
\textbf{Step 2: Derive User Persona.} Based on the image intent, conceive a reasonable user role---who would need to examine, inspect, or judge this image in their life or work? \emph{Requirements:} (i)~Be specific to a profession or role, rather than a generic ``user''; (ii)~there should be a natural causal relationship between the persona and the image intent.\\[2pt]
\textbf{Step 3: Design Judgment Task.} Based on the persona, design a complex rule-based judgment task that this role might perform in real-world settings. \emph{Requirements:}\\
- The task is a high-level description explaining ``from what aspects'' the image needs to be inspected, but does not need to list the specific rules.\\
- The task should be concise and clear, have a non-trivial difficulty level, require reference to multiple rules in order to be answered, and cannot be resolved by visual observation alone.\\
- The aspects inspected should be understandable and judgeable based on general domain common sense and world knowledge, and should not rely on specific industry professional regulations, technical standards, or domain terminology (e.g., avoid requiring knowledge of legal article numbers, medical diagnostic standards, or engineering specification parameters).\\
- The task should not degenerate into a simple yes/no classification task, but require inference combined with the image content.\\
- The task should have a real-world demand background in practical scenarios.\\
- The final judgment result should be a binary conclusion of ``compliant / non-compliant'' or ``pass / fail''.\\[4pt]
\#\# Output Format\\[2pt]
Strictly output a JSON object:\\[2pt]
\{\\
\hspace*{1em}"image\_intent":\ "<Brief analysis of image intent, 1--2 sentences>",\\
\hspace*{1em}"persona":\ "<User persona name>",\\
\hspace*{1em}"task":\ "<High-level description of the multi-criteria judgment task, explaining the aspects to be inspected and the final judgment goal, without listing specific rules>"\\
\}
}
\end{tcolorbox}
\vspace*{\fill}
\caption{Prompt used in Stage~1 to jointly extract image intent, derive a domain-specific persona, and propose a multi-criteria judgment task.}
\label{fig:persona-prompt}
\end{figure*}

\clearpage
\begin{figure*}[p]
\centering
\vspace*{\fill}
\begin{tcolorbox}[
    colback=gray!5,
    colframe=teal!75!black,
    coltitle=white,
    fonttitle=\bfseries,
    title={Prompt Used for Rubric Quality Assessment},
    boxrule=0.6pt,
    arc=2pt,
    left=6pt, right=6pt, top=4pt, bottom=4pt
]
\textbf{[SYSTEM PROMPT]}\\[2pt]
{\fontsize{8pt}{10pt}\selectfont\ttfamily
You are a multimodal data quality assessment expert. Your task is to evaluate the quality of a set of automatically generated evaluation criteria (rubrics).\\[4pt]
\#\# Input Information\\[2pt]
You will receive the following information:\\
- \textbf{Image}: An image to be judged.\\
- \textbf{Persona}: Describes the user role performing the task.\\
- \textbf{Task}: A description of a complex rule-based judgment task, requiring systematic inspection of the image based on multiple rules.\\
- \textbf{Rubrics}: A set of numbered binary judgment rules (R1, R2, ...) automatically generated by the model. Each rule specifies a pass/fail criterion and is tagged with a Type (Perceptual, Reasoning, Content, Format, or Linguistic) and a Priority level (Critical, Important, or Optional).\\[4pt]
\#\# Evaluation Dimensions\\[2pt]
Please evaluate the rubric set along the following four dimensions, providing a score of 1--5 and a brief rationale for each.\\[4pt]
\textbf{Dimension 1: Specificity.} \emph{Object: each rule.} Is each rule precisely described, with actionable judgment criteria and clear ``compliant / non-compliant'' boundaries? Are the judgment objects identifiable in the image (i.e., presence or absence can be determined)?\\[2pt]
\textbf{Dimension 2: Reasonableness.} \emph{Object: each rule.} Under the given persona and task, is each rule logically grounded and of practical inspection value? Reasonableness is judged by the persona/task, \emph{not} by whether the inspected object actually appears in the image.\\[2pt]
\textbf{Dimension 3: Completeness.} \emph{Object: the rule set.} Does the rubric cover all judgment dimensions required by the task, including those implied by the task description and realized only through image content?\\[2pt]
\textbf{Dimension 4: Coherence.} \emph{Object: the rule set.} Are the rules semantically independent, free of redundant repetition and logical contradictions, and as a whole well-aligned with the persona and task?\\[4pt]
For each dimension, use the following anchored scale: \textbf{5}~=~excellent / negligible defects; \textbf{4}~=~strong with minor issues on isolated rules; \textbf{3}~=~mixed, with several clearly problematic rules; \textbf{2}~=~majority of rules problematic on this dimension; \textbf{1}~=~near-complete failure on this dimension.\\[4pt]
\#\# Output Format\\[2pt]
Strictly output a JSON object with one entry per dimension:\\[2pt]
\{\\
\hspace*{1em}"specificity":\ \{"score":\ <1-5>,\ "rationale":\ "..."\},\\
\hspace*{1em}"reasonableness":\ \{"score":\ <1-5>,\ "rationale":\ "..."\},\\
\hspace*{1em}"completeness":\ \{"score":\ <1-5>,\ "rationale":\ "..."\},\\
\hspace*{1em}"coherence":\ \{"score":\ <1-5>,\ "rationale":\ "..."\}\\
\}
}
\end{tcolorbox}
\vspace*{\fill}
\caption{Prompt used in Stage 3 rubric quality assessment.}
\label{fig:filter-prompt}
\end{figure*}

\clearpage
\begin{figure*}[p]
\centering
\vspace*{\fill}
\begin{tcolorbox}[
    colback=orange!4,
    colframe=orange!75!black,
    coltitle=white,
    fonttitle=\bfseries,
    title={Prompt Used for Structured Response Generation},
    boxrule=0.6pt,
    arc=2pt,
    left=6pt, right=6pt, top=4pt, bottom=4pt
]
\textbf{[OUTPUT PROMPT]}\\[2pt]
{\fontsize{8pt}{10pt}\selectfont\ttfamily
[Output Requirements]:\\[4pt]
The rubrics above contain numbered rules (R1, R2, ...), each tagged \texttt{[Type|Priority]}.\\
- \textbf{Type} guides your reasoning approach: P~(Perceptual)~=~visual evidence, R~(Reasoning)~=~logical inference, C~(Content)~=~coverage check, F~(Format)~=~structural check, L~(Linguistic)~=~style/tone check.\\
- \textbf{Priority} determines weight: Critical (any fail $\to$ overall Fail), Important (major quality factor), Optional (bonus).\\[4pt]
\textbf{Think step by step:}\\[4pt]
\textbf{Step 1 --- Visual Grounding.} Identify and describe the visual elements in the image that are relevant to the task and rubrics.\\[2pt]
\textbf{Step 2 --- Rule-by-Rule Verification.} Evaluate each rule in priority order (Critical first, then Important, then Optional). For each rule $R_i$ \texttt{[Type|Priority]}:\\
\hspace*{1em}\textbullet\ Apply type-appropriate reasoning: cite visual evidence for \texttt{[P]}, perform logical inference for \texttt{[R]}, check information scope for \texttt{[C]}, verify structure for \texttt{[F]}, assess language style for \texttt{[L]}.\\
\hspace*{1em}\textbullet\ Conclude with Pass or Fail and a brief justification.\\[2pt]
\textbf{Step 3 --- Aggregation \& Conclusion.} Aggregate all rule results. If any Critical rule is Fail, overall result is Fail. Otherwise, weigh Important and Optional results. State the decisive rule(s) and the overall judgment.\\[4pt]
Strictly output a JSON object in the following format:\\[2pt]
\{\\
\hspace*{1em}"thinking\_process":\ "Step 1: ...$\backslash$nStep 2:$\backslash$n- R1 [Type|Priority]: Pass/Fail. Justification...$\backslash$n- R2 [Type|Priority]: Pass/Fail. Justification...$\backslash$n...$\backslash$nStep 3: ...",\\
\hspace*{1em}"final\_answer":\ \{\\
\hspace*{2em}"visual\_grounding":\ "Key visual elements relevant to the task",\\
\hspace*{2em}"rule\_results":\ \{"R1":\ "Pass",\ "R2":\ "Fail",\ ...\},\\
\hspace*{2em}"decisive\_rules":\ "R1, R3",\\
\hspace*{2em}"overall\_result":\ "Pass/Fail"\\
\hspace*{1em}\}\\
\}
}
\end{tcolorbox}
\vspace*{\fill}
\caption{Prompt used in Stage~4 to elicit the structured discriminative trace $\tau = (g, \{(j_i, e_i)\}, y)$. Type tags steer the per-rule reasoning mode, while priority tags impose a Critical-first verification order and an aggregation rule under priority precedence.}
\label{fig:response-prompt}
\end{figure*}

\end{document}